# RegionGCN: Spatial-Heterogeneity-Aware Graph Convolutional Networks

Hao Guo[a], Han Wang[b], Di Zhu[c], Lun Wu[a], A. Stewart Fotheringham[d] and Yu Liu[a]*

[a]*Institute of Remote Sensing and Geographical Information Systems, School of Earth and Space Sciences, Peking University, Beijing, China*

[b]*Future Urbanity & Sustainable Environment (FUSE) Lab, Division of Landscape Architecture, Faculty of Architecture, The University of Hong Kong, Pokfulam, Hong Kong Special Administrative Region, China*

[c]*Geospatial Data Intelligence (GeoDI) Lab, Department of Geography, Environment and Society, University of Minnesota, Twin Cities, Minneapolis, MN, USA*

[d]*Spatial Data Science Center, Department of Geography, Florida State University, Tallahassee, FL, USA*

*Corresponding author: Yu Liu (liuyu@urban.pku.edu.cn)

# RegionGCN: Spatial-Heterogeneity-Aware Graph Convolutional Networks

Modeling spatial heterogeneity in the data generation process is essential for understanding and predicting geographical phenomena. Despite their prevalence in geospatial tasks, neural network models usually assume spatial stationarity, which could limit their performance in the presence of spatial process heterogeneity. By allowing model parameters to vary over space, several approaches have been proposed to incorporate spatial heterogeneity into neural networks. However, current geographically weighting approaches are ineffective on graph neural networks, yielding no significant improvement in prediction accuracy. We assume the crux lies in the over-fitting risk brought by a large number of local parameters. Accordingly, we propose to model spatial process heterogeneity at the regional level rather than at the individual level, which largely reduces the number of spatially varying parameters. We further develop a heuristic optimization procedure to learn the region partition adaptively in the process of model training. Our proposed spatial-heterogeneity-aware graph convolutional network, named RegionGCN, is applied to the modeling of county-level vote share in the 2016 US presidential election based on socioeconomic attributes. Results show that RegionGCN achieves significant improvement over the basic and geographically weighted GCNs. We also offer an exploratory analysis tool for the spatial variation of non-linear relationships through ensemble learning of regional partitions from RegionGCN. Our work contributes to the practice of Geospatial Artificial Intelligence (GeoAI) in tackling spatial heterogeneity.

Keywords: spatial process heterogeneity, GeoAI, graph neural networks, graph convolutional networks, election analysis

**Introduction**

In the emerging field of Geospatial Artificial Intelligence (GeoAI) (Janowicz et al. 2020; Hu et al. 2024), there has been substantial progress in recent years in analyzing spatial data with deep learning technologies. A key advantage of these methods is their weak model assumptions (Zhu, Gao, and Cao 2022). Based on the universal approximation ability of neural networks, deep learning methods avoid the need for prior specification of the function form, spatial weights matrix, and distance decay functions in spatial models. Such methods therefore offer the capacity to model the complicated relationships between spatial variables (Zhu et al. 2020, Simini et al. 2021, Yao and Huang 2023). Rather than simply applying general neural network models, it is crucial to integrate spatial principles to better tackle geo-spatial problems (Li 2020). A spatially explicit model not only includes domain knowledge but also tends to achieve improved performance (Li et al. 2021, Xie et al. 2021).

As a crucial aspect of spatial modeling, spatial process heterogeneity causes the relationships between spatial variables (i.e. spatial process) to vary over space (Anselin 2010, Fotheringham and Sachdeva 2022). Therefore, a global model fitted for all locations would ignore important variations in the spatial processes being modeled leading to biased explanations and inferior model predictions. This presents a challenge when neural network models are applied to spatial data because, although the mapping between input features and target variables could be very complex, it is commonly assumed to be spatially stationary (Janowicz et al. 2022). A possible solution is to jointly model the global and local aspects of the spatial processes to reach a balance between generality and heterogeneity (Goodchild and Li 2021). Examples in spatial statistical modelling include Multiscale Geographically Weighted Regression (MGWR) (Fotheringham, Yang, and Kang 2017; Fotheringham, Oshan and Li, 2024) and spatial regime regression (Anselin and Amaral 2024). There have also been recent explorations in GeoAI to integrate spatial heterogeneity with neural networks, represented by GWANN (Geographically Weighted Artificial Neural Network, Hagenauer and Helbich 2022), GNNWR (Geographically Neural Network Weighted Regression, Du et al. 2020) and STAR (Spatial Transformation And moderation, Xie et al. 2021).

Due to their adaptability to model data with irregular spatial structures and explicit consideration of spatial dependency, GNNs have become a useful tool for spatial modeling (Liu and Biljecki 2022) and have been applied in various classification and prediction tasks involving traffic flow (Bai et al. 2020, Guo et al. 2022), human activity intensity (Li et al. 2021), air quality (Han et al. 2023), building patterns (Yan et al. 2019), urban function (Hu et al. 2021), and regional economies (Xu, Li, and Xu 2020). A geographically weighted graph convolutional network (GWGCN) has also been proposed (Zhu et al. 2022). However, explicitly modeling spatial process heterogeneity by introducing local parameters, as in GWGCN, exhibits inferior node-level prediction performance compared to basic GCN according to results in Zhu et al. (2022). We suspect this may be caused by introducing too many additional parameters, which can largely increase the risk of over-fitting and limit the extrapolation of results to different locations.

If this is the case, a solution to the problem is to model spatial process heterogeneity between larger regions instead of across the individual spatial units. In this way, the processes being modelled are assumed to be homogeneous within regions, and heterogeneous across regions. Such regions are referred to as spatial regimes (Anselin and Amaral 2024)[1]. We assume such a regional modeling approach will achieve a better balance between heterogeneity and generality in the context of GCNs. Compared with unit-specific local parameters, using region-specific parameters in GCN will substantially reduce the number of parameters to be estimated, which potentially addresses the over-fitting issue of GWGCN. Furthermore, it is desirable not to have to define the regional portioning of the data *a priroi* and subjectively but to adaptively learn the optimal region partition from the data (Guo, Python, and Liu 2023).

We propose RegionGCN, a graph-based machine learning framework with explicit modeling of spatial process heterogeneity at the regional level. Built upon the proposed regionally weighted graph convolution, RegionGCN captures process heterogeneity using a restricted number of additional parameters, which successfully mitigates the over-fitting risk. We also introduce a dynamic zoning module based on iterative region optimization which allows automatic learning of the regionalization scheme from the data instead of an *a priori* specification. We apply our method to county-level vote-share prediction in the 2016 US Presidential election. Results show that RegionGCN outperforms not only the basic and geographically weighted GCNs, but also a DeepWalk approach intended to enhance the input with location information. The results demonstrate the usefulness of explicitly modeling spatial process heterogeneity in graph neural networks. Through ensemble learning of multiple zoning results produced by RegionGCN, our method also provides a tool of exploration analysis into heterogeneous, complex spatial relationships.

**Related Work**

Spatial process heterogeneity and spatial data dependency are considered the two major properties that distinguish spatial analysis from aspatial analysis (Anselin 2010). When applied to spatial grids and graphs, the convolution operation in Convolutional Neural Networks (CNNs) and Graph Neural Networks (GNNs) aggregates neighborhood information, thus addressing the issue of spatial data dependency. Its flexibility is further enhanced by deformable convolution (Zhang, Yu, and Zhu 2022) and dynamic graph convolution (Guo et al. 2022). However, general neural network architectures lack consideration for the problem of spatial process heterogeneity (Janowicz et al. 2022).

In spatial analytics, models for spatial process heterogeneity roughly fall into two categories (Anselin and Amaral 2024). The local approach, exemplified by Geographically Weighted Regression (GWR) (Fotheringham, Brunsdon, and Charlton 2002) and Multiscale Geographically Weighted Regression (MGWR) (Fotheringham, Oshan and Li, 2024), allows model parameters to vary across spatial units (throughout the paper, a 'spatial unit' corresponds to an observation). The stratified approach, exampled by spatial regime regression (Guo, Python, and Liu 2023, Anselin and Amaral 2024), divides the spatial units into several groups, and model parameters are shared inside groups and different across groups. The development of spatial-heterogeneity-aware neural networks in GeoAI has been parallel with these two lines.

For the local approach, Du et al. (2020) proposed GNNWR, which is further extended to spatio-temporal modelling (Wu et al. 2021) and convolutional neural networks (Dai et al. 2022). Instead of the weighting scheme being based on a spatial kernel function as in GWR and MGWR, GNNWR utilizes a neural network to directly model the mapping from spatial positions to regression coefficients. GNNWR has shown a superior ability to model spatial process heterogeneity compared with GWR (Du et al. 2020), yet the relationship between dependent and independent variables is still assumed to be linear. Hagenauer and Helbich (2022) designed the GWANN, where the parameters of the output neurons are allowed to vary across space. The model is trained with a geographically weighted loss function, enabling the modeling of non-linear, spatially heterogeneous relationships. In the field of graph-based deep learning, Zhu et al. (2022) developed GWGCN, which introduces a set of multiplicative parameters for each spatial unit in graph convolution layers. This framework supports semi-supervised learning and is especially useful when the proportion of labeled data is low. Nevertheless, the prediction accuracy is shown to be compromised by introducing too many spatially varying parameters.

For the stratified approach, Gupta et al. (2021) proposed Spatial Variability Aware deep Neural Networks (SVANN). Their framework trains a neural network independently with data from each spatial region, and attains better total accuracy than a single model trained with the whole data set. However, two limitations exist for this framework. First, it relies on a given spatial partition scheme, which may not accord with heterogeneity of the spatial processes. Second, as the data set is split according to the *a priori* spatial partition, the amount of data available for training each regional model may be inadequate. The spatial ensemble learning framework developed by Jiang et al. (2019) addresses the first issue for classification tasks. The STAR framework proposed by Xie et al. (2021) addresses both issues in a more general manner and is applicable to both classification and regression tasks. Based on the modeling accuracy, STAR automatically partitions spatial area in a hierarchical way. The closer two spatial units are in the spatial hierarchy, the more parameters they share in the neural network. Besides neural networks, the STAR framework is also extended to tree-based models such as random forests (Xie et al. 2025). Nevertheless, the framework is still not applicable to graph neural networks, which limits its application on graph-structured spatial data. The research described here falls into the stratified approach and can be viewed as the stratified counterpart of GWGCN, as well as an extension of the idea behind STAR to graph neural networks.

**The RegionGCN Model**

We address node-level prediction tasks on a spatial graph $G = (V, E)$. Each node $v_i$ ($i = 1, \cdots, n$) in the node set $V$ corresponds to a location or a spatial areal unit which is associated with observed features $x_i$ and a target variable $y_i$ to be predicted. In this research, we consider regression tasks where the target variable takes continuous values although the model can also be applied to categorical dependent variables. The edge $e_{ij} = (v_i, v_j, w_{ij})$ in the edge set $E$ represents existence (where $w_{ij}$ takes a binary value) or intensity (where $w_{ij}$ takes non-negative real values) of the connection between $v_i$ and $v_j$. Examples of connections include geographic proximity, attribute similarity, or spatial interaction intensity (Zhu et al. 2020). We assume the input features of all locations are known, while the target variable is only sampled for some of the locations. Our goal is to predict the target variable at unsampled locations. Graph neural network models are advantageous for such tasks, as they explicitly model spatial dependency through message passing between linked nodes. Moreover, GNNs utilize features of unsampled locations during training, achieving semi-supervised learning. Here we present RegionGCN, a new graph neural network model, to enable the differentiation of heterogeneous relationships over space and address the over-fitting issue in the local weighting approach (Zhu et al. 2022).

Our model is based on a generalized form of graph convolution layer (Hamilton 2020) as follows,

$$X^{(l)} = \sigma(D^{-1}AX^{(l-1)}\Theta^{(l)} + X^{(l-1)}\Phi^{(l)} + \Psi^{(l)}) \tag{1}$$

where $X^{(l)} \in \mathbb{R}^{n \times c_l}$ is the matrix of output node embeddings of the $l$th graph convolution layer ($c_l$ is the number of embedding dimensions; in particular, $c_0$ is the number of input features), $\sigma(\cdot)$ is the non-linear activation function. $D$ and $A$ are the degree matrix and adjacency matrix, respectively, of the graph $G$; $D^{-1}A$ is the row-normalized adjacency matrix which performs an average pooling of neighborhood embeddings. Unlike common settings, two matrices of trainable weight parameters $\Theta^{(l)}, \Phi^{(l)} \in \mathbb{R}^{c_{l-1} \times c_l}$ are introduced to process the information from the neighborhood and the current node separately (Liu et al. 2020). $\Psi^{(l)} \in \mathbb{R}^{n \times c_l}$ are the matrix of trainable bias parameters. Note that $\Psi^{(l)} = (b_1 \mathbf{1}_n, \cdots, b_{c_l} \mathbf{1}_n)$ contains only $c_l$ trainable parameters. By introducing the two weight matrices, $\Theta^{(l)}$ and $\Phi^{(l)}$, the generalized graph convolution enables the model to adjust the relative importance of current-node and neighborhood information, which appears to be essential for competitive performance in our experiment. In the Supplementary Information, we discuss in more detail the rationale of adopting the generalized graph convolution instead of the basic form.

Following the idea of GWGCN (Zhu et al. 2022), a locally weighted graph convolution layer can be formulated based on Equation 1 as follows:

$$X^{(l)} = \sigma(D^{-1}A(X^{(l-1)} \odot \Omega_{\text{loc}}^{(l)})\Theta^{(l)} + (X^{(l-1)} \odot \Omega_{\text{loc}}^{(l)})\Phi^{(l)} + \Psi^{(l)}) \qquad (2)$$

where $\Omega_{\text{loc}}^{(l)} \in \mathbb{R}^{n \times c_{l-1}}$ is a matrix of trainable geographical weights, represented as $\Omega_{\text{loc}}^{(l)} = (\eta_1^{(l)}, \cdots, \eta_n^{(l)})^{\text{T}}$, where $\eta_i^{(l)} \in \mathbb{R}^{c_{l-1}}$ is the local weight vector corresponding to node $v_i, i = 1, \cdots, n$; $\odot$ represents Hadamard (element-wise) product between matrices. The local weighting operation introduces $nc_{l-1} = O(n)$ parameters in $\Omega_{\text{loc}}^{(l)}$, while the number of trainable parameters in the unweighted form (Equation 1) is $(2c_{l-1} + 1)c_l$. The number of spatial observations $n$ can be in the order of $10^3$ in GCN applications, which is much higher than the embedding dimensions $c_{l-1}$ and $c_l$ (Li et al. 2021, Zhu et al. 2022). Hence, the number of parameters in GWGCN is much larger than the unweighted GCN. This could make the GWGCN model prone to over-fitting. Moreover, for nodes in the test set, the local parameters are learned only based on the accuracy of nodes in the training set in their $L$-hop neighborhood, where $L$ is the number of graph convolution layers. This may affect the validity of these local weights.

We propose a regionally weighted graph convolution (RegConv) layer, which is based on a partition of the nodes into $p$ spatial regions, each assumed to be governed by a stationary spatial process. The role of these regions is similar to those in spatial regime regression. Generally, we do not require the regions to be geographically connected. Instead of learning specific parameters for each node as Equation 2, we introduce specific parameters for each region, thus avoiding the introduction of $O(n)$ additional network parameters. The RegConv layer is defined as follows:

$$X^{(l)} = \sigma((D^{-1}AX^{(l-1)}\Theta^{(l)} + X^{(l-1)}\Phi^{(l)}) \odot \Omega_{\text{reg}}^{(l)} + \Psi_{\text{reg}}^{(l)}) \quad (3)$$

where $\Omega_{\text{reg}}^{(l)} \in \mathbb{R}^{n \times c_l}$ and $\Psi_{\text{reg}}^{(l)} \in \mathbb{R}^{n \times c_l}$ are matrices of trainable region-specific weight and bias, respectively. Let $\boldsymbol{r} = (r_1, \cdots, r_n)$ be the allocation vector of the $n$ nodes, where $r_i \in \{1, \cdots, p\}$ is the index of the region containing node $i$, and the region $j$ has trainable weight vector $\omega_j \in \mathbb{R}^{c_l}$ and bias vector $\psi_j \in \mathbb{R}^{c_l}$. We have

$$\Omega_{\text{reg}}^{(l)} = (\omega_{r_1}^{(l)}, \cdots, \omega_{r_n}^{(l)})^{\text{T}} \quad (4)$$

$$\Psi_{\text{reg}}^{(l)} = (\psi_{r_1}^{(l)}, \cdots, \psi_{r_n}^{(l)})^{\text{T}} \quad (5)$$

Figure 1 illustrates the difference between the locally and regional weighting approaches. The number of trainable parameters in Equation 3 is $2(p + c_{l-1})c_l$, which is much less than Equation 2 given $p \ll n$.

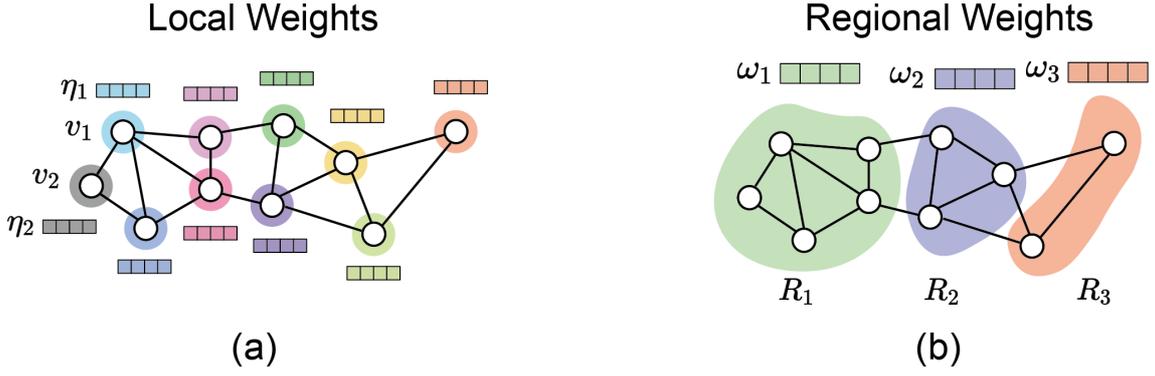

Figure 1. Schematic diagrams of (a) local weighting and (b) regional weighting on a spatial graph. $\eta_i$ represent the local weight vector corresponding to node $v_i$, and $\omega_j$ represent the regional weight vector corresponding to region $R_j$. These spatial weights operate on embeddings of spatial units through element-wise product.

As the region scheme associated with spatial process heterogeneity is often unknown, we develop a heuristic zoning procedure, achieving the joint learning of model parameters and region delineation. The algorithmic procedure is given in Algorithm 1 in the Supplementary Information.

Before training RegionGCN, the region scheme $r$ is randomly initialized using a region growth procedure similar as in Guo, Python, and Liu (2023). First, $p$ distinct nodes are randomly selected with equal probability, which are assigned as seed nodes for the $p$ regions. Then for each region $j$ ($j = 1, \cdots, p$), unassigned nodes linked to nodes in region $j$ are assigned to region $j$. This step is repeated until all the nodes are assigned to a region.

After training the model for every $T_0$ epochs ($T_0$ is a predefined hyperparameter), we reallocate nodes to improve the region delineation, following the idea of the Automatic Zoning Procedure of Openshaw and Rao (1995). A node is defined as a boundary node if at least one of its neighbors belongs to a different region. In each iteration, the algorithm tries to move each boundary node to one of its neighboring regions which yields the lowest training loss. The move is maintained if the training loss decreases, otherwise the system reverts to the previous state. Assume a node $v$ is moved from region $R_1$ to $R_2$, a lower loss indicates that the parameters of $R_2$ describe the process at $v$ better than $R_1$. Therefore, by reallocating the nodes to minimize the loss, the algorithm gathers nodes with similar process into the same region. The procedure stops when no node is moved throughout an iteration. Note that neural network parameters, including the region-specific weight and bias parameters, are frozen during the region optimization procedure[2].

A RegionGCN contains $L$ stacked RegConv layers, followed by a linear transformation layer (Figure 2). Both the global parameters $\Theta^{(l)}, \Phi^{(l)}$ and the region-specific parameters $\omega_j^{(l)}, \psi_j^{(l)}$ are trained through back-propagation, and the region partition is optimized during this process. To capture the global relationship and provide a basis for learning spatial heterogeneous effects, we first train a general GCN built with generalized graph convolution layers (Equation 1) on the same data, and the learned parameters are used to initialize RegionGCN. Details on the model training procedure is given in the Supplementary Information.

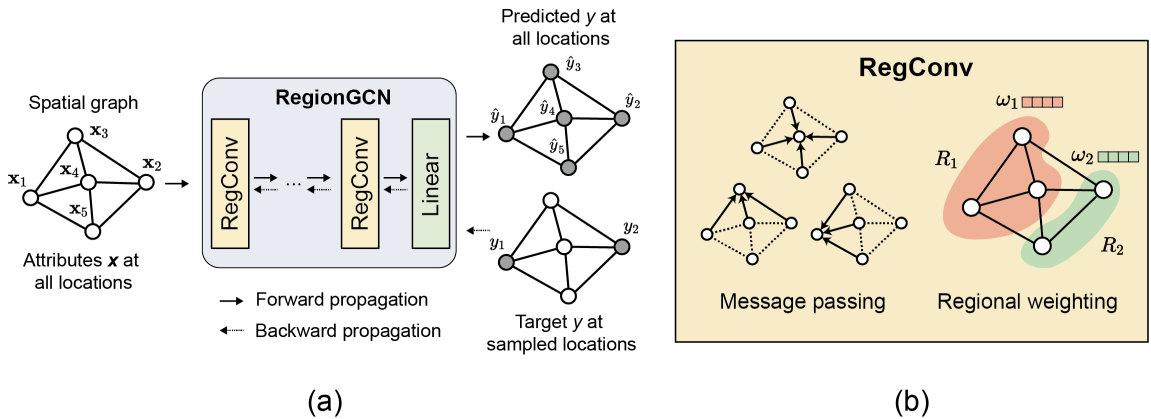

Figure 2. (a) The architecture and training schema of RegionGCN. (b) An illustration of the regionally weighted graph convolution layer (RegConv).

**Investigating Spatial Contexts with a Region Ensemble**

The regions derived by RegionGCN also provide a tool for exploratory analysis of non-linear, heterogeneous spatial processes. Nevertheless, the region scheme produced by a single run is easily affected by the split of training and test data sets, as well as by the randomness in model training. Assuming the target variable is available for all locations, we repeat the modeling process for $K$ rounds ($K$ is a predefined hyperparameter). For each round, we randomly split the data, independently train a RegionGCN model to predict the target variable, and obtain an optimal regional scheme. Note that in order to control over-fitting, we still hold out a test set rather than letting the model fit the entire dataset. We argue that the ensemble of the $K$ region schemes reflects spatial heterogeneity of the modeled relationships more reliably than does any one particular scheme. The interpretation of these regions is similar to endogenously derived spatial regimes in spatial regression (Anselin and Amaral 2024), although our method has the advantage of not having to assume the form of the modeled relationships between the variables.

To obtain the ensemble region scheme, we apply the Cluster-based Similarity Partitioning Algorithm (CSPA) proposed by Strehl and Ghosh (2002). CSPA was originally proposed to produce cluster ensembles, with no consideration of spatial contiguity. To get geographically connected regions, we follow the modification in Aydin et al. (2018), where CSPA is utilized for ensemble learning of tree-based regionalization. First, a similarity graph $G_s$ is defined. The nodes and links of $G_s$ are the same as the spatial graph $G$, except that $G_s$ is a weighted graph, and the weight of an edge linking nodes $v_i$ and $v_j$ is the frequency that $v_i$ and $v_j$ belong to the same region among the $K$ region partition results. These edge weights reflect the process similarity between nodes.

Second, a graph partitioning algorithm is applied to $G_s$, removing a subset of edges ("cut-set") and producing $R$ connected regions. $R$ is a user-defined parameter and may be different from the number of regions $p$ in RegionGCN. Following Strehl and Ghosh (2002) and Aydin et al. (2018), we use the METIS k-way partition algorithm (Karypis and Kumar 1998). The goal of this algorithm is to minimize the total edge weight of the cut-set, which produces regions with homogeneous processes in our case. Actually, an edge with a low weight connects two nodes with dissimilar processes and should be pruned to divide the nodes into different regions. The algorithm also applies a constraint on the balance of the region size, with size being defined as the number of nodes a region contains. The ratio of the maximum region size and average region size cannot exceed $1 + u/1000$ ($u$ is a user-defined parameter). However, as our purpose is to reveal the spatial pattern of process heterogeneity, the region size should be allowed to vary. We use large $u$ values to weaken this constraint.

Following Strehl and Ghosh (2002), the performance of a region ensemble is measured by Average Normalized Mutual Information (ANMI). Given two region schemes of the same set of nodes $\mathcal{R} = \{R_i\}$, $\mathcal{R}' = \{R'_j\}$ (regions are viewed as sets of containing nodes), their similarity is measured by Normalized Mutual Information (NMI, Vinh et al. 2010):

$$\text{NMI}(\mathcal{R}, \mathcal{R}') = \frac{-2\sum_i\sum_j |R_i \cap R'_j| \log\frac{n|R_i \cap R'_j|}{|R_i||R'_j|}}{\left(\sum_i |R_i| \log\frac{|R_i|}{n}\right) + \left(\sum_j |R'_j| \log\frac{|R'_j|}{n}\right)} \qquad (6)$$

Given $K$ individual region schemes $\mathcal{R}_1, \cdots, \mathcal{R}_K$ from repeated RegionGCN runs and the ensemble region scheme $\bar{\mathcal{R}}$,

$$\text{ANMI} = \frac{1}{K}\sum_{i=1}^{K} \text{NMI}(\mathcal{R}_i, \bar{\mathcal{R}}) \qquad (7)$$

ANMI takes value in [0,1], and a higher value indicates better overall consistency with individual region schemes.

**An Empirical Example: Spatial Prediction of Voting in the 2016 U.S. Presidential Election**

Analysis of peoples' election voting behavior is an important topic in political geography and political science. Previous studies have revealed spatially varying associations between socioeconomic features and voting outcome (Fotheringham, Li, and Wolf 2021). Such "spatial contextual effects" may result from variations in local customs, psychological traits, spatially varying media influence and so on (Fotheringham and Li 2023). The spatially heterogeneous nature of voting behavior makes vote share prediction suitable for demonstrating the usefulness of RegionGCN. Moreover, compared with local spatial regression approaches such as MGWR, our region ensemble approach provides a novel perspective to spatial context analysis by relaxing the linearity assumption of the relationships being modeled.

The Appendix provides another empirical example to model the county-level violent crime rate in the U. S. We use RegionGCN to impute missing values in the dataset, further demonstrating the practical value of the model's prediction capability.

*Data*

In this example, the target variable is the county-level vote share of the Democratic Party in the 2016 US Presidential election. To be specific, if $N_{\text{dem}}$, $N_{\text{rep}}$ denote the number of votes for the Democratic and Republic Party, respectively, then the target variable, *r*, is

$$r = \frac{N_{\text{dem}}}{N_{\text{dem}} + N_{\text{rep}}} \tag{8}$$

The vote share data cover $n = 3108$ counties in the Contiguous US (Figure 3). The choice of input features is aligned with a validated model in previous works (Fotheringham, Li, and Wolf 2021; Fotheringham and Li 2023). The input features include 12 socioeconomic variables covering age, gender, ethnicity, education, and economic status, along with 2 variables related to voting (Table 1). All socioeconomic data are from the 2012-2016 American Community Survey (ACS) 5-year estimates[3], and voting data are provided by MIT Election Data and Science Lab[4]. All input variables are standardized to zero mean and unit variance. A global ordinary least squares (OLS) linear regression suggests that these variables explain up to 64% of the variance in the vote share of the Democratic Party (Table 2). Regarding the spatial graph, we construct an undirected, unweighted graph based on rook's contiguity so that two counties are linked in the spatial graph if, and only if, they share a common border[5].

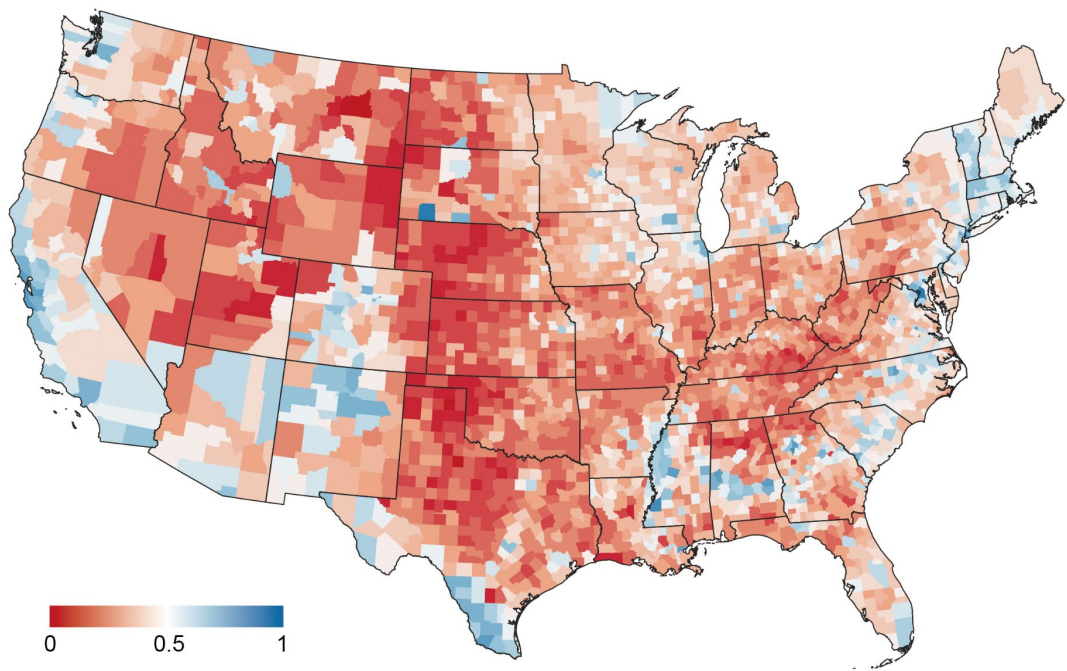

Figure 3. County-level vote share of Democratic Party among votes for Democratic and Republic Party in the 2016 US Presidential election. Blue indicates more votes for Democratic Party, red otherwise.

Table 1. The input features for vote share prediction.

| Category | Feature | Explanation |
|---|---|---|
| Population | ln(Pop_Den) | Logarithm of population per square kilometres |
| Age | Pct_Age_18_29 | Percentage of persons between 18 years and 29 years |
| | Pct_Age_65 | Percentage of persons 65 years and over |
| Gender | Sex_Ratio | Ratio between male and female population |
| Ethnicity | Pct_Black | Percentage of black or African American |
| | Pct_Hispanic | Percentage of Hispanic or Latino |
| | Pct_FB | Percentage of foreign-born population |
| Education | Pct_Bachelor | Percentage of Bachelor's degree or higher among persons age 25+ |
| Economic Status | Median_Income | Median household income in the past 12 months in dollars |
| | Gini | Gini Index of Income Inequality |
| | Pct_Insured | Percentage of health-insured population |
| | Pct_Manuf | Percentage of population employed in the manufacturing industry |
| Voting | Turnout | Voter turnout based on voting age population |
| | Pct_3rd_Party | Percentage of third-party vote |

Table 2. The OLS linear regression results for vote share

| Variables | Estimate | Standard Error | P>|t| |
|---|---|---|---|
| Intercept | 33.290* | 0.174 | 0.000 |
| ln(Pop_Den) | 4.648* | 0.274 | 0.000 |
| Pct_Age_18_29 | 0.511 | 0.286 | 0.074 |
| Pct_Age_65 | 0.527 | 0.286 | 0.065 |
| Sex_Ratio | 0.161 | 0.198 | 0.418 |
| Pct_Black | 8.406* | 0.227 | 0.000 |
| Pct_Hispanic | 4.316* | 0.291 | 0.000 |
| Pct_FB | 1.981* | 0.314 | 0.000 |
| Pct_Bachelor | 4.984* | 0.382 | 0.000 |
| Median_Income | -3.331* | 0.372 | 0.000 |
| Gini | 0.855* | 0.232 | 0.000 |
| Pct_Insured | 1.512* | 0.255 | 0.000 |
| Pct_Manuf | 0.449* | 0.219 | 0.041 |
| Turnout | 2.267* | 0.291 | 0.000 |
| Pct_3rd_Party | 3.900* | 0.224 | 0.000 |

Note: *Significant at 0.05 level. Model $R^2$=0.641. No variance inflation factors are greater than 5.

## Prediction Results

We implement a RegionGCN with two RegConv layers, and compare the vote share prediction performance of RegionGCN with the following models:

- **LR**: A linear regression model estimated with OLS. Note that this is different from the model in Table 2 as here we hold out test data. The implementation of the Python package *scikit-learn* (Pedregosa et al. 2011) is used.
- **SLX**: A spatially lagged X model (Halleck Vega and Elhorst 2015). The neighborhood average of each input feature is added to LR, yielding 28 explanatory variables altogether.
- **XGBoost**: Extreme Gradient Boosting, an ensemble learning method based on regression trees. The implementation of the Python package *xgboost* is used.
- **GWR**: A Geographically Weighted Regression model with an adaptive bisquare kernel (Fotheringham, Brunsdon, and Charlton 2002)[6]. The bandwidth is selected based on the corrected Akaike information criterion (AICc). We use the implementation of the Python package *mgwr* (Oshan et al. 2019).
- **ANN**: A fully connected neural network with no message passing among nodes.

- **GCN**: A graph convolutional neural network with no explicit modeling of spatial heterogeneity. The generalized graph convolution in Equation 1 is used.
- **GWGCN**: A geographically weighted graph neural network (Zhu et al. 2022) with node-specific parameters (Equation 2).

All the neural network models, including ANN, GCN, GWGCN, and RegionGCN, are implemented using PyTorch. Other implementation details are given in the Supplementary Information. The set of nodes is randomly partitioned into training, validation, and test sets at a ratio of 6:2:2. Only the training set labels are used for back-propagation. Labels of the validation set are used for hyperparameter tuning. Labels of the test set are used for performance evaluation only. The data partition procedure is repeated $K = 10$ times.

We report the average performance across the 10 random data splits for each model in Table 3. The basic neural network model ANN outperforms LR, SLX and XGBoost, yet is inferior to GWR. This indicates the usefulness of considering spatial heterogeneity. All graph convolutional networks outperform GWR and ANN, with the mean absolute error lower than 4%. This demonstrates the usefulness of modeling spatial dependency in voting behavior. Although we have used a large $L_2$ regularization factor to mitigate over-fitting, the prediction performance of GWGCN is still slightly inferior to GCN. This result verifies the findings in Zhu et al. (2022) that GWGCN is more suitable for data fitting rather than prediction. RegionGCN achieves the lowest error among all considered models, significantly outperforming GCN and GWGCN (both with P-value < 0.01, one-tailed paired samples t-test for RMSE, similarly hereinafter). These results demonstrate the effectiveness of introducing region-level parameters to model spatial heterogeneity and mitigate over-fitting.

Table 3. The spatial prediction accuracy of vote share for considered models. Metrics are averaged over 10 random data splits.

| Model | RMSE | MAE | $R^2$ |
|---|---|---|---|
| LR | 9.6674 | 7.0741 | 0.6388 |
| SLX | 8.4630 | 6.2674 | 0.7228 |
| XGBoost | 6.6776 | 4.8888 | 0.8280 |
| GWR | 5.7571 | 4.0967 | 0.8719 |
| ANN | 6.5826 | 4.8545 | 0.8319 |
| GCN | 5.2081 | 3.8795 | 0.8952 |
| GWGCN | 5.3303 | 3.9633 | 0.8902 |
| RegionGCN | **4.9888** | **3.6903** | **0.9039** |

Note: RMSE: Root Mean Squared Error; MAE: Mean Absolute Error; $R^2$: the coefficient of determination. The best value of each metric is put in bold.

***Comparison with Model Variants***

We compare three variants of RegionGCN to investigate the effectiveness of the region optimization procedure, and the necessity of the region contiguity constraint (which is not considered in RegionGCN):

- **RegionGCN-R**: RegionGCN with fixed, random region partition. The regions are not adaptively updated to align with the spatial heterogeneous process. Instead, the randomly generated regions are used throughout the training of RegionGCN.
- **RegionGCN-P**: RegionGCN with fixed, pre-clustered region partition. The regions are defined as feature-based clusters from the K-Means algorithm[8], and not adaptively updated while training RegionGCN.

- **RegionGCN-C**: RegionGCN with contiguous region partition. Spatial contiguity is maintained during the region optimization procedure. If re-assignment of a unit would break region contiguity, the re-assignment is cancelled[7].

Results in Table 4 show that the prediction accuracy of RegionGCN-R and RegionGCN-P is significantly inferior to RegionGCN (P-value < 0.01) and close to GCN. This result demonstrates that the adaptive region optimization module is essential for accurate spatial prediction, and cannot be simply replaced with feature-based clustering. The ineffectiveness of feature-based clustering in our task is related to the distinction between spatial data heterogeneity and spatial process heterogeneity (Fotheringham and Sachdeva 2022). The performance of RegionGCN-C is also inferior to RegionGCN according to all three metrics, with higher RMSE values significant at a P-value of 0.02. This result indicates that requiring the regions to be contiguous could compromise the prediction performance of RegionGCN, as such a constraint prohibits some possible moves to improve the region partition. Therefore, we do not apply the spatial contiguity constraint by default.

Table 4. The spatial prediction accuracy of vote share for variants of RegionGCN. Metrics are averaged over 10 random data splits.

| Model | RMSE | MAE | $R^2$ |
|---|---|---|---|
| RegionGCN | **4.9888** | **3.6903** | **0.9039** |
| RegionGCN-R | 5.1810 | 3.8719 | 0.8963 |
| RegionGCN-P | 5.2168 | 3.8985 | 0.8949 |
| RegionGCN-C | 5.0390 | 3.7420 | 0.9019 |

Note: RMSE: Root Mean Squared Error; MAE: Mean Absolute Error; $R^2$: the coefficient of determination. The best value of each metric is put in bold.

*Incorporating DeepWalk Embeddings*

Node embedding approaches (Hamilton 2020) extract the positional and structural information in the graph, and produce a set of additional node features (i.e. embeddings) which could be concatenated with the attribute features to enhance the prediction performance. When nodes in the graph represent spatial locations, these approaches augment the model input with spatial information, which is essentially modeling spatial heterogeneity from a different perspective from our regional weighting. Thus, it is important to test whether regional weighting is still useful when such node embeddings are incorporated.

In this section, we consider DeepWalk (Perozzi, Al-Rfou, and Skiena 2014), a prominent node embedding approach based on random walks. Starting from each node, multiple node sequences are sampled via unbiased random walks on the graph. The similarity of the embeddings of any two nodes (measured by Euclidean distance in the embedding space) is aligned with the frequency that they appear in the same node sequence. Note that DeepWalk does not use node attributes, but only the links between nodes. We use the DeepWalk implementation in PyG (PyTorch Geometric) to extract embeddings of US counties in our spatial graph. Each node embedding vector has 14 dimensions. The extracted node embeddings are visualized using t-SNE (t-distributed Stochastic Neighbor Embedding, van der Maaten and Hinton 2008) in Figure 4. t-SNE is capable of preserving the local structures in the high-dimensional embedding space when reducing the data into two dimensions. After 50 epochs, counties of the same state begin to cluster, and the distribution of clusters resembles the real map of US states (Figure 4a). The DeepWalk loss converges after 100 epochs. Although the pattern appears more distorted, the spatial proximity is largely maintained (Figure 4b). Hence, we may view the DeepWalk embeddings as a form of "coordinates" on the graph.

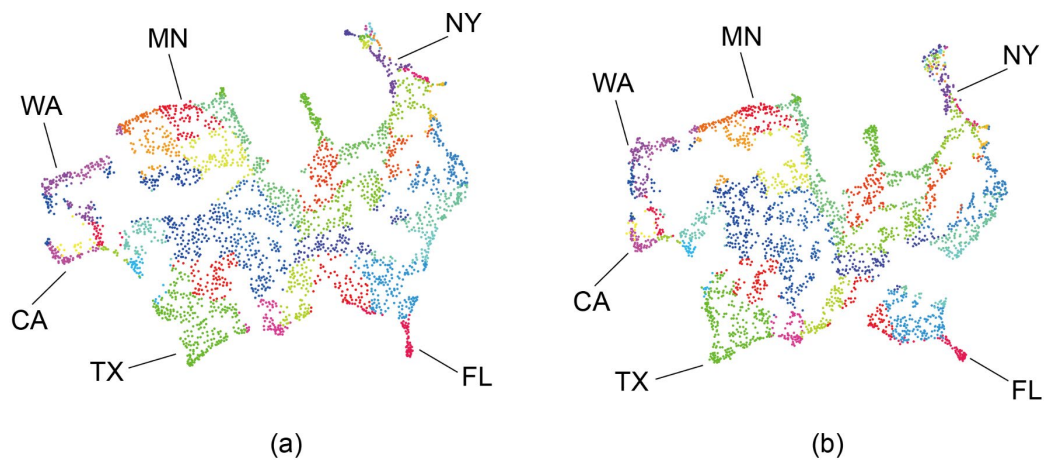

Figure 4. t-SNE visualization of DeepWalk embeddings on the connectivity graph of US counties after (a) 50 epochs; (b) 100 epochs. Each point represents a county, and colors correspond to 48 states and District of Columbia in the Contiguous US. Both subfigures have been reoriented to align with the US geography.

We concatenate the DeepWalk embeddings with node features described in Table 1, resulting in a 28-dimensional input for each node. Such enhanced features are used as model input for our vote share prediction task. We compare three neural network models in this section: ANN, GCN, and RegionGCN. The hyperparameters used are given in Table S1 in the Supplementary Information, and the accuracy metrics are in Table 5. When DeepWalk embeddings are used, RegionGCN still achieves the lowest error, outperforming ANN and GCN. Furthermore, the accuracy of GCN-DeepWalk is inferior to RegionGCN without DeepWalk embeddings according to the RMSE values (P-value＜0.01) – although RegionGCN has a slightly higher MAE. These results further demonstrate the usefulness of the proposed regional weighting approach. Compared with models using attribute input only, DeepWalk embeddings significantly improve the performance of ANN (P-value< 0.01) and GCN (P-value= 0.03), yet have little effect on the performance of RegionGCN. We infer that the spatial process heterogeneity effects captured by regional weighting and DeepWalk have some overlap, which limits the further enhancement of model performance.

Table 5. The spatial prediction error of vote share for neural network models enhanced with DeepWalk embeddings. Metrics are averaged over 10 random data splits.

| Model | RMSE | MAE | $R^2$ |
| --- | --- | --- | --- |
| ANN-DeepWalk | 5.6478 | 4.0348 | 0.8766 |
| GCN-DeepWalk | 5.0707 | 3.6718 | 0.9005 |
| RegionGCN-DeepWalk | **4.9892** | **3.6018** | **0.9037** |

Note: RMSE: Root Mean Squared Error; MAE: Mean Absolute Error; $R^2$: the coefficient of determination. The best value of each metric is put in bold.

*Region Ensemble*

Finally, we apply the region ensemble approach. Ten region schemes are used to build the similarity graph, each from training RegionGCN on a random-split dataset. We set $u = 6000$ in the graph partitioning algorithm[9]. A range of region counts $R$ from 5 to 50 (with step 5) are tried, and $R = 50$ is used as indicated by the best ANMI value of 0.458. In our result, the actual number of regions, 44, is lower than the target value as empty regions are generated in the graph partitioning process.

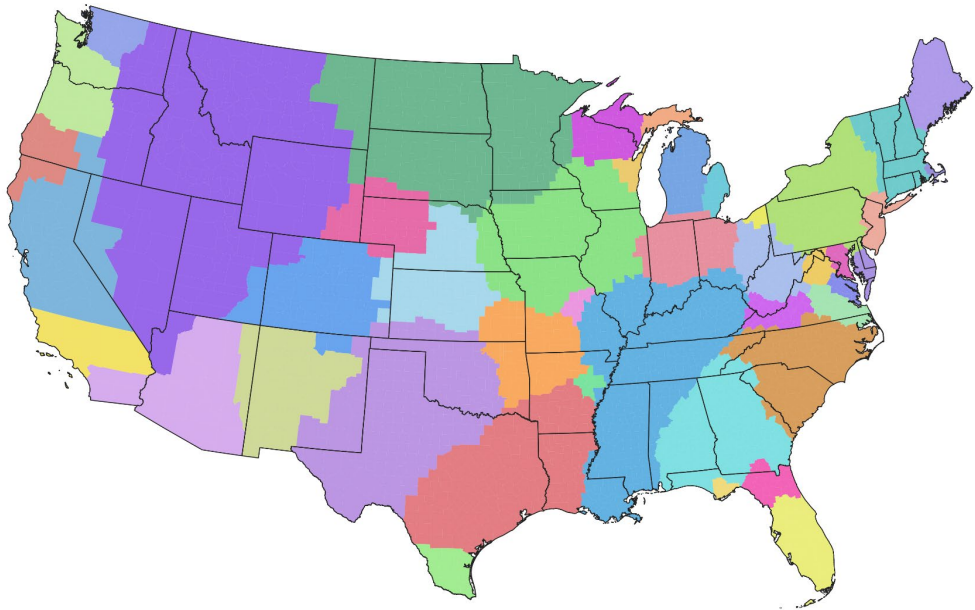

Figure 5. The map of spatial context in the 2016 US election. RegionGCN partitions on 10 random data splits are used to produce the ensemble regions.

An important part of the output from this analysis is the regionalization of the country into areas where the processes affecting voter preferences are relatively stable internally and different externally, as shown in Figure 5. In some cases, these regions, depicting where the influences on voter preference are relatively stable, coincide with state boundaries (even though state boundaries are not part of the data used in the analysis). For instance, the boundaries of South and North Carolina (combined), New Jersey (together with the Long Island), Maine, and Colorado are close to internally consistent regions within which the processes determining voting behaviour are very similar. Equally, the boundaries between Arkansas and Mississippi and between South Dakota and Nebraska can also be seen quite clearly. Moreover, within-state heterogeneity is clear in some cases. For instance, in California, two coastal regions can be identified in terms of different processes affecting voting behaviour: one in Northern California centered on San Francisco; the other in Southern California centered on Los Angeles. Similarly, Virginia is partitioned into six sub-regions indicating important variations in the processes affecting voting behaviour across the state; Texas is divided into northwestern and southeastern regions and Washington is divided into three parts with western Washington aligning with Oregon in terms of the determinants of voting behaviour. These findings largely resonate with our understanding of voting behaviour.

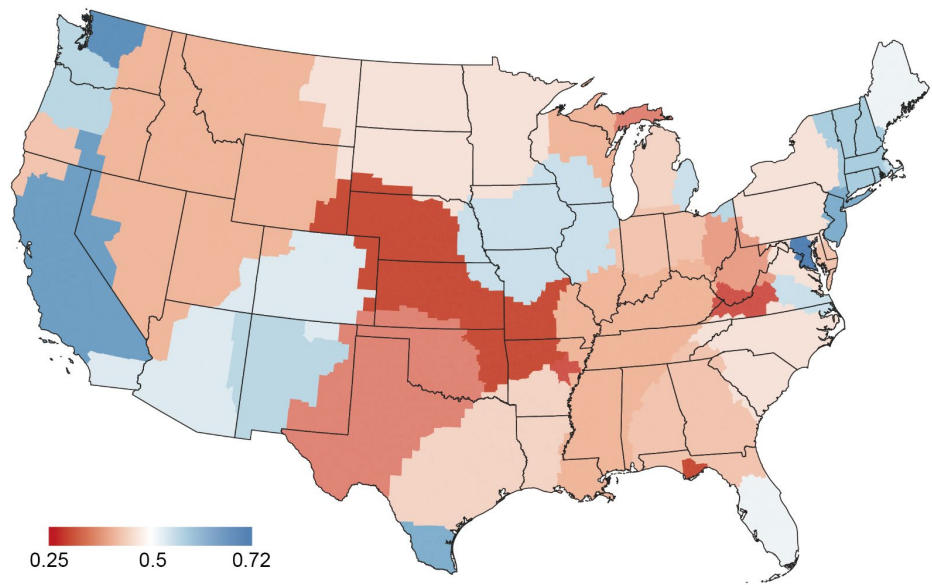

Figure 6. The vote share of Democratic Party of each region in the ensemble result. Blue indicates more votes for Democratic Party, red otherwise.

The disparities in spatial context also lead to the regions identified in Figure 5 having different voting outcomes, as shown in Figure 6 where shades of blue indicate a preference for the Democratic party and shades of red indicate a preference for the Republican party. There is a clear preference for the Democratic Party along the west coast, in the North-East and in New Mexico and southern Texas, while there is a strong preference for the Republic Party in northwestern Texas, Nebraska, much of the South and the southern part of the Mid-west.

**Conclusions and Discussion**

This research demonstrates that spatial-heterogeneity-aware graph neural networks can not only provide a better model of the processes generating a variable of interest but also achieve better spatial prediction performance on unobserved data. The key to their success is to introduce regional weights rather than local weights, which effectively restricts the number of trainable neural network parameters, especially when the number of nodes is large. Moreover, we find that regional weighting outperforms random-walk based embeddings in the modeling of spatially heterogeneous processes. These two approaches reflect different strategies in spatially explicit machine learning, where spatial elements are incorporated into the model design (Xie, Jia, et al. 2021, Zhu et al. 2022) and the input features (Li 2022), respectively. Our result supports the former approach, shedding light on the potential of developing spatially-aware models in GeoAI practice.

A major limitation of RegionGCN lies in its computational cost. Due to the large solution space (Keane 1977), the iterative approach to optimize the region scheme can be time-consuming. This problem is not serious in spatial regime regression based on linear models (Guo, Python, and Liu 2023). Yet for RegionGCN, a forward pass is needed every time when the algorithm tries to change the region assignment of a node. As a consequence, the training time of RegionGCN is about a magnitude longer than that of GCN in our analysis. Currently, RegionGCN is not scalable to large spatial datasets with over $10^4$ spatial units, according to our experience. A possible solution is to use fuzzy regions, where each node may be associated with multiple regions. For each region, the degrees of membership of each node may be trained by back-propagation simultaneously with other neural network parameters. Such an idea is reflected in Node Adaptive Parameter Learning (NAPL)[10] for traffic forecasting (Bai et al. 2020). Nevertheless, the model interpretability is compromised as the regions become fuzzy. On the other hand, extending the idea behind RegionGCN to time-series analysis is another direction for future research. For example, by interpreting the changing process of spatial heterogeneity along a time series of house price, we will be able to track the evolving structure of housing market.

The proposed region ensemble approach can still be sensitive to the randomness in data split and neural network training. Hence, the resulting region scheme is exploratory rather than confirmative. Our analysis only considers the METIS k-way partition algorithm to partition the similarity graph. Future work should investigate the application of other graph partitioning methods. Furthermore, spatial contextual effects contain intrinsic and behavioral effects (Fotheringham and Li 2023), which correspond to the spatially varying intercept and slope coefficients in a linear model, respectively. The ensemble region scheme from RegionGCN may reflect a mixture of these two effects. Although the region-specific weight and bias parameters resemble the intercept and slopes of a linear model, their distinction is blurred due to the stacked convolution operation. Moreover, the spatial weights in RegionGCN may be replaced with learnable latent features. Hence, the relationship of latent variables and spatially varying parameters in the context of GeoAI is worthy of investigation. Finally, although RegionGCN could identify the region partition associated with spatial heterogeneity, the black-box nature of neural networks makes it hard to interpret the difference in the underlying spatial processes. Future research may apply explainable artificial intelligence (XAI) techniques to RegionGCN (Liu, Zhang, and Biljecki 2024). For example, mapping each input feature's spatially varying contribution to model prediction would provide an analogy of the local parameter maps from MGWR, while also relaxing the latter's assumption on function forms. Such advances in model explainability may provide deeper insights into heterogeneous spatial phenomena.

# Notes

1. Anselin and Amaral (2024) requires spatial regimes to be geographically connected regions. In this research, spatial regimes are groups of spatial units, whether connected or not, each corresponding to a specific set of model parameters.

2. The space of possible allocation vectors (region schemes) is finite (the size is $p^n$). After each move, the training loss strictly decreases. Hence the procedure is guaranteed to stop in finite steps.

3. The population density is calculated based on the population from 2012-2016 ACS 5-year estimate, and the area provided in county boundary shapefiles in 2016 from Census.gov.

4. Available at https://doi.org/10.7910/DVN/VOQCHQ. The voter turnout is calculated based on the population over 18 from 2012-2016 ACS 5-year estimate.

5. To make the graph connected, we add links to island counties based on shipping lines. In Washington State, San Juan County is linked to Skagit County. In Massachusetts State, Nantucket County, Dukes County, and Barnstable County are linked to each other.

6. Generally, MGWR should be the default GWR. Here we do not use MGWR because the *mgwr* package has not yet supported spatial prediction with MGWR.

7. For a unit $e$ with only one neighbor $f$ (e.g. when $e$ is completely surrounded by $f$), re-assigning $f$ almost always breaks region contiguity (except when the region contains only $e, f$ themselves). Hence, we consider $e, f$ as a whole when re-assigning units.

8. The number of clusters in K-Means is equal to the fine-tuned number of regions $p$ in RegionGCN (30 in this case). The fourteen explanatory variables are used for clustering.

9. In our experiment, setting $u$ to even larger values would lead to inferior ANMI values, or cause problems from trying to bisect an empty subgraph.

10. The degree membership is represented as an $n \times p$ trainable parameter matrix. Hence, NAPL still introduces $O(n)$ parameters. Although NAPL achieves good performance on time-series prediction, it may face an over-fitting risk similar to GWGCN when applied to cross-sectional data.

**Funding**

This work was supported by the National Natural Science Foundation of China under Grant No. 42430106.

**Acknowledgments**

We thank Xuechen Wang and Yi Wang for suggestions on model design; Weiyu Zhang for discussion on model training and visualization; Yuanqiao Hou for discussion on model training; and Ziyi Wang for discussion on result interpretation.

**References**

Anselin, L. 2010. Thirty years of spatial econometrics. *Papers in Regional Science* 89(1): 3–25.

Anselin, L., and P. Amaral. 2024. Endogenous spatial regimes. *Journal of Geographical Systems* 26(2): 209–234.

Aydin, O., M. V. Janikas, R. Assunção, and T.-H. Lee. 2018. SKATER-CON: Unsupervised regionalization via stochastic tree partitioning within a consensus framework using random spanning trees. In *GeoAI'18: Proceedings of the 2nd ACM SIGSPATIAL International Workshop on AI for Geographic Knowledge Discovery*, ed. Y. Hu, S. Gao, S. Newsam, and D. Lunga, 33–42. New York: ACM.

Bai, L., L. Yao, C. Li, X. Wang, and C. Wang. 2020. Adaptive graph convolutional recurrent network for traffic forecasting. *Advances in Neural Information Processing Systems* 33: 17804–17815.

Dai, Z., S. Wu, Y. Wang, H. Zhou, F. Zhang, B. Huang, and Z. Du. 2022. Geographically convolutional neural network weighted regression: a method for modeling spatially non-stationary relationships based on a global spatial proximity grid. *International Journal of Geographical Information Science* 36(11): 2248–2269.

Du, Z., Z. Wang, S. Wu, F. Zhang, and R. Liu. Geographically neural network weighted regression for the accurate estimation of spatial non-stationarity. *International Journal of Geographical Information Science* 34(7): 1353–1377.

Fotheringham, A. S., C. Brunsdon, and M. Charlton. 2002. *Geographically Weighted Regression: The Analysis of Spatially Varying Relationships*. London: Sage Publications.

Fotheringham, A.S., and Z. Li. 2023. Measuring the unmeasurable: Models of geographical context. *Annals of the American Association of Geographers* 113(10): 2269–2286.

Fotheringham, A.S., Z. Li, and L. J. Wolf. 2021. Scale, context, and heterogeneity: A spatial analytical perspective on the 2016 U.S. presidential election. *Annals of the American Association of Geographers* 111(6): 1602–1621.

Fotheringham, A. S., and M. Sachdeva. 2022. Modelling spatial processes in quantitative human geography. *Annals of GIS* 28(1): 5–14.

Fotheringham, A. S., T. M. Oshan and Z. Li. 2024. *Multiscale Geographically Weighted Regression*. Baton Rouge: CRC Press.

Fotheringham, A.S., W. Yang, and W. Kang. 2017. Multiscale geographically weighted regression (MGWR). *Annals of the American Association of Geographers* 107(6): 1247–1265.

Goodchild, M.F., and W. Li. 2021. Replication across space and time must be weak in the social and environmental sciences. *Proceedings of the National Academy of Sciences of the United States of America* 118(35): e2015759118.

Guo, H., A. Python., and Y. Liu. 2023. Extending regionalization algorithms to explore spatial process heterogeneity. *International Journal of Geographical Information Science* 37(11): 2319–2344.

Guo, S., Y. Lin, H. Wan, X. Li, and G. Cong. 2022. Learning dynamics and heterogeneity of spatial-temporal graph data for traffic forecasting. *IEEE Transactions on Knowledge and Data Engineering* 34(11): 5415–5428.

Gupta, J., C. Molnar, Y. Xie, J. Knight, and S. Shekhar. 2021. Spatial variability aware deep neural networks (SVANN): A general approach. *ACM Transactions on Intelligent Systems and Technology* 12(6): 76.

Hagenauer, J., and M. Helbich. 2022. A geographically weighted artificial neural network. *International Journal of Geographical Information Science* 36(2): 215–235.

Hamilton, W. L. 2020. *Graph Representation Learning*. Cham: Springer Nature Switzerland.

Han, J., H. Liu, H. Xiong, J. Yang. 2023. Semi-supervised air quality forecasting via self-supervised hierarchical graph neural network. *IEEE Transactions on Knowledge and Data Engineering* 35(5): 5230–5243.

Halleck Vega, S., and J. P. Elhorst. 2015, The SLX Model. *Journal of Regional Science* 55(3): 339-363.

Hu, S., S. Gao, L. Wu, Y. Xu, Z. Zhang, H. Cui, and X. Gong. 2021. Urban function classification at road segment level using taxi trajectory data: A graph convolutional neural network approach. *Computers, Environment and Urban Systems* 87: 101619.

Hu, Y., M. Goodchild, A.-X. Zhu, M. Yuan, O. Aydin, B. Bhaduri, S. Gao, W. Li, D. Lunga, and S. Newsam. 2024. A five-year milestone: reflections on advances and limitations in GeoAI research. *Annals of GIS* 30(1): 1–14.

Janowicz, K., S. Gao, G. McKenzie, Y. Hu, and B. Bhaduri. 2020. GeoAI: spatially explicit artificial intelligence techniques for geographic knowledge discovery and beyond. *International Journal of Geographical Information Science* 34(4): 625–636.

Janowicz, K., R. Zhu, J. Verstegen, G. McKenzie, B. Martins, and L. Cai. 2022. Six GIScience ideas that must die. *AGILE GIScience Series*, 3: 7.

Jiang, Z., A. M. Sainju, Y. Li, S. Shekhar, and J. Knight. 2019. Spatial ensemble learning for heterogeneous geographic data with class ambiguity. *ACM Transactions on Intelligent Systems and Technology* 10(4): 43.

Karypis, G., and V. Kumar. 1998. A fast and high quality multilevel scheme for partitioning irregular graphs. *SIAM Journal on Scientific Computing* 20(1): 359–392.

Keane, M. 1977. The size of the region-building problem. *Environment and Planning A* 7(5): 575-577.

Li, M., S. Gao, F. Lu, K. Liu, H. Zhang, and W. Tu. 2021. Prediction of human activity intensity using the interactions in physical and social spaces through graph convolutional networks. *International Journal of Geographical Information Science* 35(12): 2489–2516.

Li, W. 2020. GeoAI: Where machine learning and big data converge in GIScience. *Journal of Spatial Information Science* 20: 71–77.

Li, Z. 2022. Extracting spatial effects from machine learning model using local interpretation method: An example of SHAP and XGBoost. *Computers, Environment and Urban Systems* 96: 101845.

Liu, P., and F. Biljecki. 2022. A review of spatially-explicit GeoAI applications in urban geography. *International Journal of Applied Earth Observation and Geoinformation* 112: 102936.

Liu, P., Y. Zhang, and F. Biljecki. 2024. Explainable spatially explicit geospatial artificial intelligence in urban analytics. *Environment and Planning B: Urban Analytics and City Science* 51(5): 1104–1123.

Liu, Z., F. Miranda, W. Xiong, J. Yang, Q. Wang, and C. T. Silva. 2020. Learning geo-contextual embeddings for commuting flow prediction. *Proceedings of the AAAI Conference on Artificial Intelligence* 34: 808–816.

Lu. H., L. Liu, H. Zhong, B. Jiang. 2024. A dose of nature to reduce sexual crimes in public outdoor spaces: Proposing the Landscape-Sexual Crime Model. Landscape and Urban Planning 250: 105143.

Openshaw, S., and L. Rao. 1995. Algorithms for reengineering 1991 census geography. *Environment and Planning A* 27(3): 425–446.

Oshan, T., Z. Li, W. Kang, L. J. Wolf, and A. S. Fotheringham. 2019. MGWR: A Python Implementation of Multiscale Geographically Weighted Regression for Investigating Process Spatial Heterogeneity and Scale. *ISPRS International Journal of Geo-Information* 8(6): 269.

Pedregosa, F., G. Varoquaux, A. Gramfort, V. Michel, B. Thirion, O. Grisel, M. Blondel, P. Prettenhofer, R. Weiss, V. Dubourg, et al. 2011. Scikit-learn: Machine learning in Python. *Journal of Machine Learning Research* 12: 2825–2830.

Perozzi, B., R. Al-Rfou, and S. Skiena. 2014. Deepwalk: online learning of social representations. In *Proceedings of the 20th ACM SIGKDD International Conference on Knowledge Discovery and Data Mining*, 701–710. New York, ACM.

Simini, F., G. Barlacchi, M. Luca, and L. Pappalardo. 2021. A deep gravity model for mobility flows generation. *Nature Communications* 12: 6576.

Strehl, A., and J. Ghosh. 2002. Cluster ensembles: A knowledge reuse framework for combining multiple partitions. *Journal of Machine Learning Research* 3: 583–617.

United States Department of Justice. Federal Bureau of Investigation. 2019. Uniform Crime Reporting Program Data: County-Level Detailed Arrest and Offense Data, United States, 2016. doi:10.3886/ICPSR37059.v3

van der Maaten, L., and G. Hinton. 2008. Visualizing data using t-SNE. *Journal of Machine Learning Research* 9(86): 2579–2605.

Vinh, N.X., J. Epps, and J. Bailey. 2010. Information theoretic measures for clusterings comparison: Variants, properties, normalization and correction for chance. *Journal of Machine Learning Research* 11(95): 2837–2854.

Wang, L., J. Yang, S. Wu, L. Hu, Y. Ge, Z. Du. 2024. Enhancing mineral prospectivity mapping with geospatial artificial intelligence: A geographically neural network-weighted logistic regression approach. International Journal of Applied Earth Observation and Geoinformation 128: 103746.

Wu, S., Z. Wang, Z. Du, B. Huang, F. Zhang, and R. Liu. 2021. Geographically and temporally neural network weighted regression for modeling spatiotemporal non-stationary relationships. *International Journal of Geographical Information Science* 35(3): 582–608.

Xie, Y., E. He, X. Jia, H. Bao, X. Zhou, R. Ghosh, and P. Ravirathinam. 2021. A statistically-guided deep network transformation and moderation framework for data with spatial heterogeneity. In *ICDM'21: Proceedings of 21$^{th}$ IEEE International Conference on Data Mining*, ed. J. Bailey, P. Miettinen, Y. S. Koh, et al., 767-776. New York: IEEE.

Xie, Y., A. N. Nhu, X.-P. Song, X. Jia, S. Skakun, H. Li, and Z. Wang. 2025. Accounting for spatial variability with geo-aware random forest: A case study for US major crop mapping. *Remote Sensing of Environment* 319: 114585.

Xu, F., Y. Li, and S. Xu. 2020. Attentional multi-graph convolutional network for regional economy prediction with open migration data. In *KDD '20: Proceedings of the 26th ACM SIGKDD International Conference on Knowledge Discovery & Data Mining*, 2225–2233. New York: ACM.

Yan, X., T. Ai, M. Yang, and H. Yin. 2019. A graph convolutional neural network for classification of building patterns using spatial vector data. *ISPRS Journal of Photogrammetry and Remote Sensing* 150: 259–273.

Yao, S., and B. Huang. 2023. Spatiotemporal interpolation using graph neural network. *Annals of the American Association of Geographers* 113(8): 1856-1877.

Zhang, Y., W. Yu, and D. Zhu. 2022. Terrain feature-aware deep learning network for digital elevation model superresolution. *ISPRS Journal of Photogrammetry and Remote Sensing* 189: 143–162.

Zhu, D., S. Gao, and G. Cao. 2022. Towards the intelligent era of spatial analysis and modeling. In *Proceedings of the 5th ACM SIGSPATIAL International Workshop on AI for Geographic Knowledge Discovery*, ed. B. Martins, D. Lunga, S. Gao, et al., 10–13. New York: ACM.

Zhu, D., Y. Liu, X. Yao, and M. M. Fischer. 2022. Spatial regression graph convolutional neural networks: A deep learning paradigm for spatial multivariate distributions. *GeoInformatica* 26(4): 645–676.

Zhu, D., F. Zhang, S. Wang, Y. Wang, X. Cheng, Z. Huang, and Y. Liu. 2020. Understanding Place Characteristics in Geographic Contexts through Graph

Convolutional Neural Networks. *Annals of the American Association of Geographers* 110(2): 408–420.

**Appendix**

We apply RegionGCN to the spatial imputation of county-level violent crime rate in the U.S. Violent crimes, including murders, rapes, robberies, and aggravated assaults, remain a serious threat to society. Accurate crime rate data are crucial for the management of police resources, and the policy making for public safety. However, such data can be incomplete at the county level, as local police may sometimes fail to report the crime data. Spatial prediction is suitable for handling this issue by imputing missing values at some locations based on observations at other locations.

In this example, the target variable is the number of violent crimes per 100,000 persons in 2016, provided by the Uniform Crime Reporting (UCR) Program (United States Department of Justice. Federal Bureau of Investigation 2019). Our analysis covers 3,108 counties in the Contiguous U.S. Besides the missing values, we filter out counties with a covered population less than 5000 to mitigate small number bias, leaving 2651 records for model training and testing. The input features include 11 socioeconomic variables from the 2012-2016 ACS 5-year estimates, representing major factors of violent crimes including socioeconomic disadvantage, ethnic distribution, unstable social conditions, gender, and age (Lu et al. 2024). The variable definition, details on data preprocessing, and model settings are given in the Supplementary Information.

The test set accuracy of considered models, averaged over 10 random data splits, is shown in Table 6. RegionGCN achieves superior prediction accuracy compared with ANN, GCN, and GWGCN. Moreover, the improvement over GCN, the second-best model, is significant at a P-value < 0.01.

The spatial prediction performance of graph convolutional networks is dependent on the choice of the spatial graph (Zhu et al. 2020). Here we explore the application of RegionGCN on flow networks. While geographic contiguity is suitable to represent neighborhood influence on political leanings, population flows may better characterize the spatial dependency between places in the crime case. We define a weighted spatial graph using 2011-2015 ACS county-level commuting flow data. GNN models using this graph are denoted as GCN-SI, GWGCN-SI, and RegionGCN-SI. Results in Table 6 show a slight improvement over the contiguity graph for GCN-SI and RegionGCN-SI, though the improvement is significant only at 0.1 level.

Table 6. The spatial prediction accuracy of violent crime rate for neural network models. Metrics are averaged over 10 random data splits.

| Model | RMSE | MAE | $R^2$ |
|---|---|---|---|
| ANN | 169.26 | 121.07 | 0.3538 |
| GCN | 162.82 | 116.03 | 0.4027 |
| GWGCN | 173.86 | 118.27 | 0.3197 |
| RegionGCN | 157.76 | 111.32 | 0.4383 |
| GCN-SI | 161.16 | 114.35 | 0.4151 |
| GWGCN-SI | 175.02 | 120.05 | 0.3107 |
| RegionGCN-SI | **155.91** | **109.73** | **0.4515** |

Note: RMSE: Root Mean Squared Error; MAE: Mean Absolute Error; $R^2$: the coefficient of determination. The best value of each metric is put in bold.

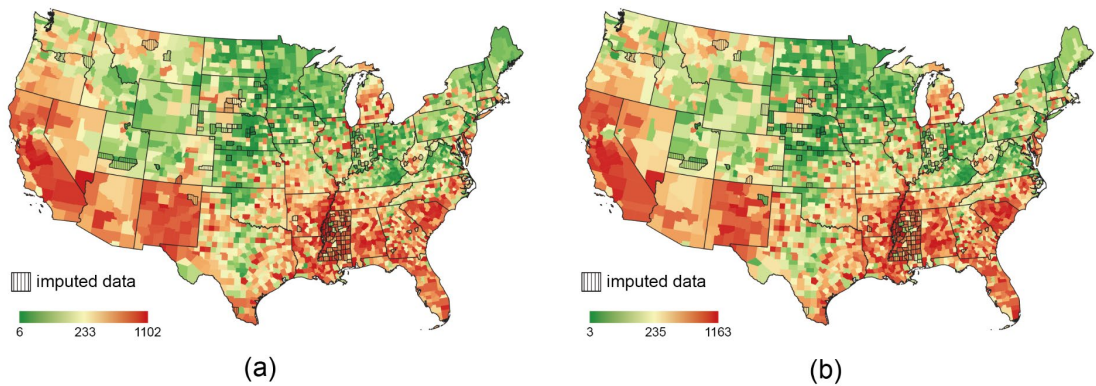

(a) (b)

Figure 7. The predicted violent crime rate (per 100,000 persons) from (a) RegionGCN; (b) RegionGCN-SI. The median values over 10 repeated runs are shown. Counties with missing values in the UCR dataset are marked with vertical lines.

Figure 7 shows the predicted county-level violent crime rates from RegionGCN and RegionGCN-SI. Both results indicate higher crime rates in the South region and southwest areas including California, New Mexico, and Arizona. This pattern is consistent with the ground truth (see Fig. S2 in the Supplementary Information). More importantly, our models produce violent crime rate estimates for the counties with missing data. For example, the crime rate for more than half of the counties in Mississippi is missing in the dataset. Among these counties, we find that Coahoma, Leflore, Washington, Holmes, and Jefferson might have high violent crime rates, as the median of imputed values over 10 repeated runs is above 500 according to both models.

# Supplementary Information

## RegionGCN: Spatial-Heterogeneity-Aware Graph Convolutional Networks

**A. Discussion on the Graph Convolution Operation for Spatial Modelling**

Let $G$ be the input graph with $n$ nodes, the basic graph convolution operation in GCN (Kipf and Welling 2017) is as follows:

$$X^{(l)} = \sigma(LX^{(l-1)}\Theta^{(l)} + \Psi^{(l)}) \tag{S1}$$

where $L \in \mathbb{R}^{n \times n}$ is the renormalized graph Laplacian:

$$L = \tilde{D}^{-1/2}\tilde{A}\tilde{D}^{-1/2} \tag{S2}$$

where $\tilde{A} = A + I$, $A$ is the adjacency matrix of the graph $G$; $\tilde{D}$ is the degree matrix calculated from $\tilde{A}$. Other notations are consistent with Equation 1.

The basic graph convolution defined above is commonly used in GeoAI literature (Zhu et al. 2020, Li et al. 2021, Zhu et al. 2022). However, the relative importance of current-node and neighborhood information is fixed in its aggregation process. Especially, when $G$ is an unweighted graph (e.g. determined by contiguity of areal units or $k$ nearest neighbors), $A$ is a binary matrix. By applying $\tilde{A} = A + I$, the information of each neighboring node is equally weighted with information of the current node, which possibly undervalues the role of the current node for node-level prediction. Indeed, GCN tends to generate similar representations for linked nodes, hence it is more suitable when linked nodes tend to be similar (Xiao et al. 2023), such as social networks. Yet linked nodes may have distinct attributes in spatial graphs. In this case, the potential over-smoothing effect of GCN can lead to deteriorated performance on node-level prediction.

The correspondence of GCN and spatially lagged regression is demonstrated in Zhu et al. (2022). Nevertheless, the relative importance of current-node and neighborhood information is adaptive in spatially lagged regression. For example, in the Spatially Lagged X (SLX) model,

$$\boldsymbol{y} = X\boldsymbol{\beta} + WX\boldsymbol{\delta} + \boldsymbol{\varepsilon} \tag{S3}$$

where $X$ is the matrix of explanatory variables, $\boldsymbol{y}$ is the vector of the dependent variable, $W$ is the spatial weight matrix with zero diagonal elements, and $\boldsymbol{\varepsilon}$ is the vector of independent errors. In this model, the effect of $X$ and $WX$ is captured by coefficients $\boldsymbol{\beta}$ and $\boldsymbol{\delta}$, respectively. If the spatial dependency effect is weak, the model would undervalue the neighborhood information by using small absolute values for $\boldsymbol{\delta}$.

The generalized graph convolution defined in Equation 1 allows similar learning of "weights" between current-node and neighborhood information. Such separate modeling of current-node information is seen in Liu et al. (2020). Our experiment shows that using the generalized form instead of the basic form can be essential for certain spatial prediction scenarios. For example, the performance of GCN based on Equation A1 is even inferior to a vanilla ANN on the 2016 US election data.

**B. RegionGCN Training Details**

To describe the training procedure of RegionGCN, we take the case of $L = 2$ as an example:

$$X^{(1)} = \sigma((D^{-1}AX^{(0)}\Theta^{(1)} + X^{(0)}\Phi^{(1)}) \odot \Omega_{reg}^{(1)} + \Psi_{reg}^{(1)})$$

$$X^{(2)} = \sigma((D^{-1}AX^{(1)}\Theta^{(2)} + X^{(1)}\Phi^{(2)}) \odot \Omega_{reg}^{(2)} + \Psi_{reg}^{(2)})$$

$$\hat{\boldsymbol{y}} = \tau(X^{(2)}\boldsymbol{u} + b\boldsymbol{1}) \tag{S4}$$

where $X = X^{(0)} \in \mathbb{R}^{n \times c_0}$ is the input node features, $\hat{y} \in \mathbb{R}^n$ is the predicted values of the target variable. In the linear transformation layer, $\boldsymbol{u} \in \mathbb{R}^{c_2}$ and $b \in \mathbb{R}$ are the weight and bias parameters, respectively; $\tau(\cdot)$ is a transformation function which may be different from $\sigma(\cdot)$.

The RegionGCN model is trained in a two-stage scheme. In Stage 1, a GCN with graph convolution layers defined in Equation 1 is trained to capture the global relationship between the features $X$ and target $y$. Region partition is not considered, nor are the region-specific parameters introduced. This stage provides a base model for learning the region partition and spatially varying parameters in RegionGCN. After Stage 1, the learned parameters $\Theta^{(l)}, \Phi^{(l)}, \Psi^{(l)}$ are used to initialize a RegionGCN model with convolution layers defined in Equation 3. To be specific, $\Theta^{(l)}, \Phi^{(l)}$ are directly copied; all the elements in $\omega_j^{(l)}, j = 1, \cdots, p$ are initialized as 1; let $\Psi^{(l)} = (\psi^{(l)}, \cdots, \psi^{(l)})^{\mathrm{T}}$, then $\psi_j^{(l)} = \psi^{(l)}, j = 1, \cdots, p$. This RegionGCN model is trained in Stage 2 to capture the spatial variation of the modeled relationship. In Stage 2, the region-specific weight and bias parameters are trained simultaneously with other network parameters through back-propagation. During this process, the region optimization procedure described in Algorithm 1 is performed to update the allocation vector $\boldsymbol{r}$.

Both stages use the mean squared error (MSE) loss and the Adam optimizer (Kingma and Ba 2015) to train the graph neural network models. The learning rate, $L_2$ regularization factor, as well as the number of regions $p$ are tuned based on validation error. An early stopping strategy is used in both stages to mitigate over-fitting. If the validation error is above the historical minimum for $T$ consecutive epochs ($T$ is a predefined hyperparameter), the training is stopped, and the model parameters at the historical minimum are maintained.

```
Algorithm 1 The region optimization procedure in RegionGCN
Input: Input features X; adjacency matrix A; training labels y_tr; current region labels r = (r_1,...,r_n); a
RegionGCN model M(X, A, r) with parameters frozen.
Output: Updated region labels r.
 1: stable = False;
 2: need_check[i] = True for i = 1,...n;
 3: while not stable do
 4:     stable = True;
 5:     for i = 1, 2, ··· , n do
 6:         if node i is a boundary node and need_check[i]=True then
 7:             Compute the training loss L_min with y_tr and predicted values from M(X, A, r);
 8:             r_min = r_i;
 9:             C = {r_k| node k is adjacent to node i} − {r_i};
10:             for j in C do
11:                 r' = (r_1,..., r_{i−1}, j, r_{i+1},..., r_n);
12:                 Compute the training loss L with y_tr and predicted values from M(X, A, r');
13:                 if L < L_min then
14:                     L_min = L; r_min = j;
15:                 end if
16:             end for
17:             need_check[i] = False;
18:             if r_i ≠ r_min then
19:                 r_i = r_min;
20:                 stable = False;
21:                 for k in {k| node k is adjacent to node i} do
22:                     need_check[k] = True;
23:                 end for
24:             end if
25:         end if
26:     end for
27: end while
28: return r.
```

## C. Supplementary Information on the U.S. Presidential Election Case Study

*C.1 Model Settings*

In the example of the 2016 U.S. Presidential Election, a RegionGCN model with two RegConv layers (as defined in Equation S4) is used, where the input dimension $c_0 = 14$ and the embedding dimension of the RegConv layers $c_1 = c_2 = 56$. We do not use a deeper network as more stacked graph convolution layers can make the node embeddings too similar (over-smoothing), which makes it difficult for the model to distinguish between nodes and make precise predictions (Hamilton 2020). The embedding dimension of 56 is enough for effective performance in practice, which is also of the same order as similar applications in GeoAI (Li et al. 2021, Zhu et al. 2022). The Rectified Linear Unit (ReLU) activation function is used for the two RegConv layers, and the sigmoid function is used in the linear output layer to transform the prediction into [0,1]. The early-stopping tolerance is $T = 1000$ epochs for the general GCN (Stage 1), and $T = 20$ epochs for RegionGCN (Stage 2). The interval of the region optimization procedure is $T_0 = 10$ epochs. The choices of $T$ and $T_0$ are trade-offs between prediction performance and computational cost.

The compared ANN, GCN, and GWGCN models all contain two hidden layers and a linear output layer, and the number of neurons in each hidden layer is aligned with RegionGCN. The fine-tuned hyperparameters for the neural network models are given in Table S1. The $L_2$ regularization factor (also called the weight decay parameter in the Adam optimizer) is set to large values for graph convolutional networks to mitigate over-fitting. Similar to RegionGCN, the early-stopping strategy is used when training the other neural network models, with the early-stopping tolerance $T = 1000$ epochs. Note that SLX and the graph convolutional networks exploited neighborhood information, while XGBoost do not. Hence, we add the neighborhood average of input features in XGBoost for fair comparison (in the spatial graph, the neighborhood of a node $v$ is defined as all the nodes that directly link to $v$). As a result, the number of input features is doubled to 28 for XGBoost. For XGBoost, the fine-tuned learning rate is 0.03, and the number of estimators is 500. As LR and SLX do not contain hyperparameters, we combine the training and validation data to train these two models.

For the models enhanced with DeepWalk embeddings, we use the PyG node2vec implementation, where DeepWalk is implemented as a special case of node2vec. We set walk_length=20, context_size=10, walks_per_node=20, and use default values for the other parameters. The embeddings are optimized using SparseAdam with a learning rate of 0.01.

Table S1. Neural network hyperparameters used in the U.S. election vote share prediction

| Model | Learning rate | $L_2$ regularization factor | Number of regions |
|---|---|---|---|
| ANN | 1e-2 | 3e-3 | / |
| GCN | 1e-3 | 0.3 | / |
| GWGCN | 1e-4 | 0.1 | / |
| RegionGCN | 1e-4 | 1.0 | 30 |
| ANN-DeepWalk | 3e-3 | 3e-4 | / |
| GCN-DeepWalk | 3e-2 | 0.1 | / |
| RegionGCN-DeepWalk | 3e-4 | 3e-2 | 30 |

*C.2 Sensitivity analysis on the number of regions*

To assess the sensitivity of model performance on the number of regions $p$, we consider a range of $p$ values from 10 to 100, and repeat the vote share prediction experiment with RegionGCN. The average performance on test data over 10 random data splits for each $p$ value is shown in Figure S1. Note that we set $p = 30$ in the main experiment according to the validation error, which may not produce the best test accuracy. As regions become smaller when $p$ increases, the generalization of region-level parameters may be affected. However, both RMSE and $R^2$ remain stable when $p \geq 20$. One possible explanation is that the other model parameters are shared across all regions, which mitigates the effect of large $p$ on region-level parameters. Moreover, during the adaptive zoning step, an empty region may appear if the region is redundant. On the other hand, the accuracy is slightly harmed when $p = 10$, indicating 10 regions are not enough to express the spatial process heterogeneity in this case.

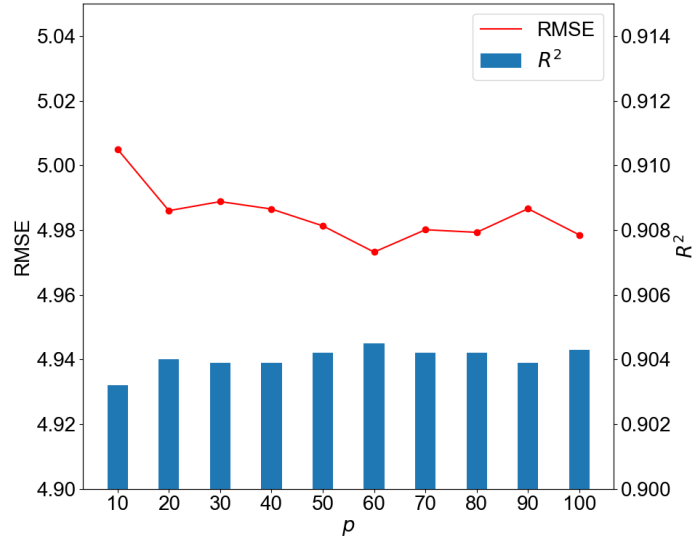

Figure S1. The vote share prediction accuracy of RegionGCN using different number of regions $p$. Average RMSE and R-squared over 10 random data splits are shown for each $p$ value.

### D. Supplementary Information on the Crime Rate Prediction

The county-level violent crime rate is calculated from the violent crime counts and covered population in Part 4 "Crimes Reported" of the UCR dataset. The data were collected from monthly reports voluntarily submitted by state, county, and city police agencies. Data completeness is measured with a coverage indicator (COVIND), which is defined as the average reporting rate (the number of months reported divided by 12) over all agencies in a county, weighted by the covered population of each agency. Among all the 3108 counties in the Contiguous U. S., 111 counties have COVIND=0 or missing covered population, so that the crime rate is unavailable; 43 more counties have COVIND<50, yielding large uncertainty for the crime rate. Both cases are considered as missing data in this study. Within the rest of records, 303 counties with covered population less than 5,000 and excluded, which may lead to small number bias in crime rate. The remaining 2,651 records, as showed in Figure S2, are randomly split to training, validation, and test data at 6:2:2. The data partition procedure is repeated 10 times. Note that we still use the complete graph structure and explanatory variables for all 3,108 counties in the model. Explanatory variables are defined in Table S2, and the OLS linear regression result is given in Table S3.

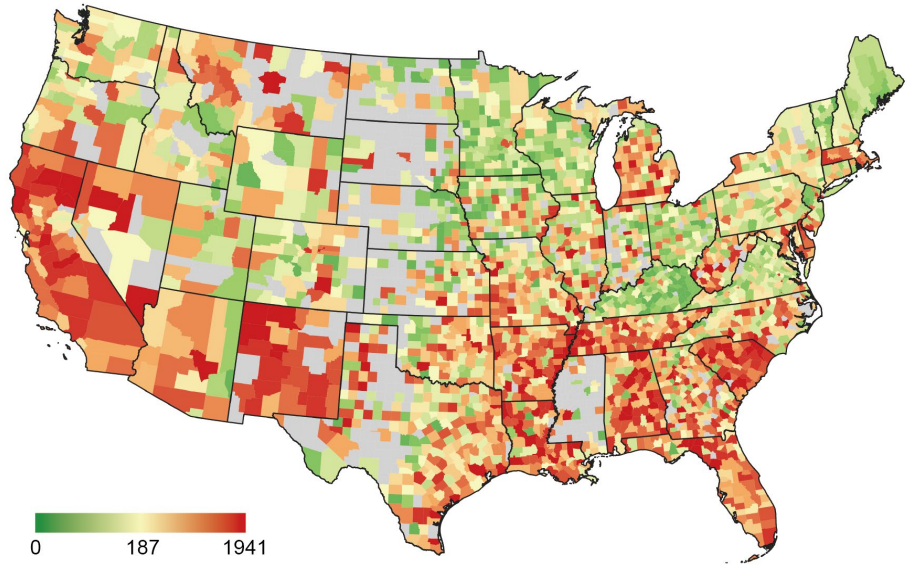

Figure S2. County-level violent crime rates in 2016 based on the UCR dataset. Gray indicates missing data, or excluded data due to small population.

Table S2. The input features for violent crime rate prediction.

| Factor | Feature | Explanation |
|---|---|---|
| Socioeconomic Disadvantage | ln(Pop_Den) | Logarithm of population per square kilometres |
| | Pct_Poverty | Percentage of persons below poverty level |
| | Pct_Unemployed | Unemployment rate among persons 16 years and over |
| | Pct_Bachelor | Percentage of Bachelor's degree or higher among persons age 25+ |
| Ethnic Distribution | Pct_Black | Percentage of black or African American |
| | Pct_Hispanic | Percentage of Hispanic or Latino |
| | Pct_Asian | Percentage of Asian |
| Unstable Social Conditions | Pct_Rent | Percentage of renter-occupied housing units among occupied housing units |
| | Pct_Vacant | Percentage of occupied housing units |
| Gender & Age | Sex_ratio | Percentage of renter-occupied housing units |
| | Pct_Age_15_34 | Percentage of persons between 15 years and 34 years |

Table S3. The OLS linear regression results for violent crime rate

| Variables | Estimate | Standard Error | P>\|t\| |
|---|---|---|---|
| Intercept | 271.309 | 3.251 | 0.000 |
| ln(Pop_Den) | 34.181 | 4.721 | 0.000 |
| Pct_Poverty | 8.909 | 5.741 | 0.121 |
| Pct_Unemployed | 3.701 | 4.657 | 0.427 |
| Pct_Bachelor | -24.100 | 5.368 | 0.000 |
| Pct_Black | 59.679 | 3.955 | 0.000 |
| Pct_Hispanic | 24.296 | 3.547 | 0.000 |
| Pct_Asian | -3.107 | 4.461 | 0.486 |
| Pct_Rent | 54.345 | 5.124 | 0.000 |
| Pct_Vacant | 13.694 | 4.076 | 0.001 |
| Sex_ratio | -14.625 | 3.802 | 0.000 |
| Pct_Age_15_34 | -6.934 | 5.091 | 0.173 |

Note: *Significant at 0.05 level. Model $R^2$=0.316. No variance inflation factors are greater than 5.

We use a RegionGCN with two hidden layers (Equation S4) where the input dimension $c_0 = 11$ and the embedding dimension of the RegConv layers $c_1 = c_2 = 44$. The ReLU activation function is used for the two RegConv layers and the linear output layer. The values of $T, T_0$ is the same as described in Section C.1. The structure of ANN, GCN, and GWGCN are aligned with RegionGCN in a similar manner with the vote share example. The fine-tuned hyperparameters based on validation error is given in Table S4.

For GCN-SI, GWGCN-SI, and RegionGCN-SI, the spatial graph is defined as the commuting flow between counties. The weight matrix is symmetrized by taking the average flow volumes of the two directions.

Table S4. Neural network hyperparameters used in the violent crime rate prediction

| Model | Learning rate | $L_2$ regularization factor | Number of regions |
|---|---|---|---|
| ANN | 0.1 | 0.01 | / |
| GCN | 1e-3 | 0.1 | / |
| GWGCN | 1e-3 | 3e4 | / |
| RegionGCN | 3e-3 | 1.0 | 30 |

**References**

Kingma, D. P., and J. L. Ba. 2015. Adam: A method for stochastic optimization. Paper presented at the 3th International Conference on Learning Representations, San Diego, CA, USA, May 7-9.

Kipf, T. N., and M. Welling. 2017. Semi-supervised classification with graph convolutional networks. Paper presented at the 5th International Conference on Learning Representations, Toulon, France, April 24-26.

Xiao, C., J. Zhou, J. Huang, T. Xu, and H. Xiong. 2023. Spatial heterophily aware graph neural networks. In *Proceedings of the 29th ACM SIGKDD Conference on Knowledge Discovery and Data Mining*, 2752–2763. New York, ACM.